Full length article

# The Outline of Deception: Physical Adversarial Attacks on Traffic Signs Using Edge Patches

*Haojie Ji[a,b], Te Hu[a,b*], Haowen Li[c], Long Jin[a,b], Chongshi Xin[a,b], Yuchi Yao[a,b], Jiarui Xiao[a,b]*

[a]*Key Laboratory of Modern Measurement & Control Technology, Ministry of Education, Beijing Information Science & Technology University, Beijing, 100192, China*
[b]*School of Mechatronical Engineering, Beijing Information Science & Technology University, Beijing, 100192, China*
[c]*Hong Kong Polytechnic University, Hong Kong, 999077, China*

A B S T R A C T

Physical adversarial attacks pose significant threats to intelligent transportation systems, especially in traffic sign recognition. Current methods often struggle to balance attack effectiveness, stealth, and transferability in real-world scenarios. In this paper, we introduce TSEP-Attack, a novel adversarial patch method that significantly enhances visual stealth and attack effectiveness. By harnessing instance segmentation and a U-Net-based generator, our approach generates adversarial patches that are finely tuned to the edge contours of traffic signs. These patches are strategically applied to the peripheral regions of the signs, ensuring minimal perceptibility to human vision. To optimize both the effectiveness and stealthiness of the patches, we design a multi-level loss function that integrates color consistency, texture matching, and frequency domain alignment, ensuring the generated perturbations seamlessly integrate with the sign's background while maintaining robustness in the physical world. Experimental results demonstrate that TSEP-Attack achieves superior attack effectiveness and stealth under limited query budgets, achieving a top success rate of up to 90%, while also exhibiting outstanding transferability across different models. Additionally, our approach maintains robust performance across varying viewing angles and distances in real-world settings, making it a viable solution for autonomous driving systems that rely on traffic sign recognition.

## 1. Introduction

Autonomous driving systems fundamentally depend on intelligent perception systems to acquire accurate environmental data, which is essential for precise vehicle control and decision-making. Among various perception modalities, vision-based technologies assume particular importance in information-rich environments due to their unique capacity to deliver comprehensive semantic understanding. Critical visual tasks include detecting and recognizing vehicles, pedestrians, traffic signs, and other essential objects, along with performing instance segmentation of roadway infrastructure. However, these vision systems demonstrate significant vulnerability to security threats, where any compromise of visual perception can severely impact both driving decisions and communication processes. When perception systems are compromised—especially within Vehicle-to-Everything (V2X) environments—erroneous visual information can lead to incorrect driving decisions and disrupt vehicle-infrastructure communication. Such security breaches ultimately undermine traffic safety and overall system efficiency, highlighting the urgent need for robust protective mechanisms in autonomous driving architectures. These visual tasks primarily rely on the powerful feature extraction capabilities of deep neural networks (DNNs), as exemplified by Tesla's Full Self-Driving (FSD) system [1], which utilizes a purely vision-based approach with deep learning to achieve end-to-end autonomous driving. This reliance on DNNs makes vision-based autonomous driving susceptible to misclassification errors, which could potentially result in severe safety hazards. However, numerous studies have demonstrated that even well-trained visual DNN models are highly vulnerable to adversarial examples. These are inputs with small, often imperceptible modifications that are crafted to mislead the model. Adversarial attacks were first prominently reported in 2013, showing that carefully crafted perturbations could mislead image classifiers [2]. Subsequent research has extensively explored such attacks across various visual tasks, including image classification [3], object detection [4], object tracking [5], semantic segmentation [6], and visual recognition [7], particularly within autonomous driving applications. These studies highlight the harmful effects of adversarial attacks, demonstrating that they can significantly degrade target model performance. Moreover, the widespread vulnerability of DNNs to such attacks poses a substantial challenge to the robustness and reliability of downstream applications.

As one of the main tasks in autonomous driving, traffic sign recognition (TSR) will directly affect the safety of autonomous driving [8]. Relying on the road information captured by cameras, TSR automatically recognizes various traffic signs by using DNNs, providing key rules and road condition information for autonomous vehicles. DNNs [9,10,11,12] have demonstrated remarkable accuracy and robustness in TSR, but the vulnerability to adversarial attacks in DNNs presents a significant security loophole in autonomous driving. The security of TSR for intelligent transportation systems is indisputable, and misclassification of traffic signs could lead to intolerable risks. However, even the most advanced traffic sign models are inevitably prone to misjudgment when confronted with adversarial samples [13]. In real-world driving scenarios, attackers can introduce perturbations to traffic signs through methods such as patch placement [14] and optical projection [15], thereby interfering with the visual information captured by cameras. Physical adversarial attacks are precisely the form of attacks that introduce adversarial perturbations into network models via physical media. The focus of such attacks is on how adversarial perturbations can be effectively input into sensors through physical means [16]such as by directly applying trained patch patterns [17] to the target sign, using semi-transparent stickers [18,19] to cover the camera, and employing projectors or lasers [20] to project light onto traffic signs.

From the perspective of effectiveness against physical adversaries, high efficacy, repeatability, low cost, and stealthiness are the key metrics in maximizing the impact of adversarial attacks. Although several studies have explored adversarial attack methods targeting TSR models, most of them have presented certain limitations. Firstly, many adversarial attack studies are conducted on image classification tasks and subsequently adapted to TSR, resulting in limited specificity in real-world physical scenarios. Secondly, physical attacks often require pre-deployment, which means the attack will be exposed for a long time in the traffic environment, presenting a significant challenge to its stealthiness. Traffic sign occlusion is hard to remain unnoticed by road users over long periods. Additionally, optical-based attacks impose stringent requirements on the driving environmental and involve substantial deployment costs, which significantly hinder their large-scale replication.

Considering the limitations of physical adversaries, this paper proposes a patch attack method targeting the edges of traffic signs, designed to align with the characteristics of TSR. The patch-based attacks typically focus on regions that attract human attention by interfering with the main shapes of traffic signs, such as speed limit numbers and directional arrows. Such attacks tend to focus on the structural features of traffic sign patterns, which aligns with drivers' instinctive visual focus when rapidly acquiring traffic information. When these key elements are obstructed or altered, the resulting visual anomalies become immediately apparent to attentive drivers, thus potentially compromising road safety. In this study, instance segmentation is performed on images from the large-scale real-world traffic sign dataset TSRD [21] using Segment Anything Model(SAM) [22] to generate specialized adversarial masks that fit the edges of traffic signs. A U-Net [23] generator is employed to learn the mapping from the latent space to adversarial patches, while multi-level constraints ensure the generated patches possess visual stealthiness and robustness in physical attacks. The main contributions of this work are as follows:

- This study proposes TSEP-Attack, a novel adversarial patch method for traffic sign recognition tasks. We utilize instance segmentation to obtain traffic sign edge masks and employ a U-Net generator with VGG19[24] natural texture priors to generate adversarial patches. Through a multi-level loss function incorporating color, texture, and frequency domain constraints, we ensure the generated patches naturally blend with the background visually, achieving effective concealment from human perception.
- We validated the effectiveness of white-box attacks on four traffic sign recognition network architectures and examined the transferability of adversarial samples across different models. Comparative experiments on the TSRD dataset demonstrated TSEP-Attack's superior performance in both attack effectiveness and stealthiness compared to three existing adversarial attack methods. Experimental results confirm that our training approach - incorporating color-difference based gating mechanism and warmup-based dynamic scheduler - enables rapid achievement of high attack success rates while maintaining excellent concealment. The generated adversarial samples exhibit remarkable transferability across different model architectures.
- The physical effectiveness of our method was verified through printed adversarial samples collected from varying distances and viewing angles in real-world environments. These tests confirm that our attacks maintain consistent effectiveness across different perspectives and distances while preserving certain transfer attack capabilities.

The remainder of this paper is organized as follows. Section 2 reviews existing research on TSR and related adversarial attack methods. Section 3 details the proposed TSEP-Attack methodology. Section 4 presents the experimental setup and results, focusing on validating attack performance across diverse models and scenarios, along with a multidimensional analysis of attack outcomes. Finally, Section 5 concludes the paper by summarizing key findings, discussing their implications for TSR systems, and outlining promising directions for future research.

## 2. Related work

*2.1. TSR*

TSR is formally defined as the process of using camera sensors to capture traffic sign images and facilitate decision-making. It is widely applied in autonomous driving to improve driving safety and comfort. As a multi-class classification problem with imbalanced class frequencies, TSR is typically divided into two primary tasks: 1) Traffic sign detection (TSD), which involves locating traffic signs within the input image; 2) Traffic sign classification (TSC), which entails predicting the category label for the detected signs. At present, TSR mainly relies on DNNs[13], with convolutional neural networks (CNNs)[22] demonstrating particularly strong performance.

Although traffic sign standards differ across countries, they generally exhibit specific shapes and colors, which has contributed to the widespread adoption of CNNs as the predominant approach for TSR. Sermanet et al.[25] first applied a multi-scale CNN (MS-CNN) without handcrafted features, surpassing human-level performance on a TSR benchmark. Subsequently, numerous CNN-based recognition models [26,27] have been developed to enhance traffic sign detection accuracy and robustness. In addition, researchers have successfully adapted general object detection models, such as the YOLO[28] series, to TSR, demonstrating both high detection speeds and competitive performance in applications.

*2.2. Adversarial attacks*

Adversarial attacks are typically classified into white-box and black-box attacks, depending on the level prior knowledge the attackers possesses regarding the target models. White-box attacks are performed with full knowledge of the target model's network architecture and typically compute adversarial loss using gradient-based backpropagation techniques. Common white-box attack methods include the fast gradient sign method (FGSM) [29], basic iterative method (BIM) [30], projected gradient descent (PGD) [31], DeepFool (DF) [32], and Carlini & Wagner (CW) [33] attacks. In contrast, black-box attacks are conducted without access to the internal information or architecture of the target model. Attackers must rely solely on the output predictions for generating attacks. Therefore, black-box attacks better represent real-world application scenarios. In such attacks, the attacker relies solely on the model's predicted labels or confidence scores to reconstruct the decision boundary and build a surrogate model. This process requires the surrogate model to sufficiently approximate the target model's predictive behavior [34]. The adversarial samples are then generated against this surrogate model. Initially, Papernot et al.[35] proposed a method for constructing adversarial samples using a locally trained surrogate model. This approach leveraged the observation that adversarial samples exhibited a certain degree of transferability, meaning that adversarial examples generated for a specific target model under a white-box attack may still be effective when applied to other models. However, transfer-based methods suffer from inefficiency and poor reliability, which has motivated the development of gradient estimation methods [36,37,38], and random search-based approaches [39,40,41]. White-box attacks target a specific model, while black-box attacks are conducted against the surrogate of the target model. Therefore, white-box attacks tend to be more direct and efficient in generating adversarial samples. Furthermore attacks that transfer adversarial perturbations from a white-box scenario to other models are commonly classified as gray-box attacks. In fact, there is currently no widely accepted explanation for the effectiveness of adversarial perturbations.

More attention needs to be paid to the context in which attacks are implemented, which can be classified into digital attacks and physical attacks. The key distinction is that digital attacks involve direct modification of the input image to the target model, while physical attacks require alterations in the real-world environment that are subsequently captured by sensors. The process of physical adversarial attacks generally involves four steps: 1) adversarial perturbation generation; 2) adversarial medium manufacturing; 3) threat image capturing; 4) attacking[16]. Physical adversarial attacks typically require substantially larger perturbation magnitudes than their digital counterparts, as the perturbations must overcome real-world sensing limitations to remain effective when captured by imaging systems. Additionally, physical adversarial attacks often necessitate more significant perturbations due to various physical constraints and environmental factors, such as spatial deformations and variations in illumination conditions.

The majority of research on adversarial attacks has been conducted on well-known image classification datasets, such as CIFAR-10[42], and ImageNet[43]. Traffic sign classification plays a crucial role in enabling autonomous driving systems to interpret traffic environments and make appropriate driving decisions. Therefore, evaluating the reliability has become a key focus in this domain. Consequently, adversarial attacks against TSR have received extensive attention over the past decade. Adversarial attack methods that were originally validated on image classification datasets are often successfully applied to traffic sign datasets, including GTSRB[44] and LISA[45] datasets. Additionally, adversarial attack methods specifically developed for traffic signs have also attracted considerable research interest. Eykholt et al.[14] were the first to investigate the robustness of traffic sign classifiers and proposed the robust physical perturbations (RP2) framework. By modeling physically plausible perturbation distributions and maximizing classification errors, they implemented adversarial perturbations in the physical world through printed black-and-white stickers and affixed them to traffic signs, effectively misleading the recognition system.

*2.3. Physical attacks on TSR*

These types of attacks involve strategic embedding of carefully designed patterns or textures into specific regions of the input image to construct the attack, thus referred to as adversarial patches. Early studies conducted on classification datasets, such as Google AP[46] and LaVAN [47], demonstrated that interference in local regions is not only feasible but also capable of being implemented in the physical world. This led to the identification of two fundamental objectives for patch attacks, e.g., maximizing the attack capability of the patch region and selecting the optimal location for embedding the patch. Without restricting patch content while maximizing adversarial perturbation, attackers can succeed with smaller patch areas and flexible placement on target objects. The strategic selection of patch location[48] further contributes to enhancing adversarial performance and minimize the required patch size[41]. In extreme scenarios, iterative optimization techniques can identify the most effective attack region, enabling the deception of recognition systems even through the modification of a single pixel[49]. Additionally, the research on patch shapes has explored which alters the minimal possible area to achieve successful adversarial effects[50].

In contrast to the global perturbation patterns typically employed in digital attacks, patch-based approaches concentrate adversarial modifications within localized regions, significantly improving their practical feasibility for real-world deployment. For traffic sign attacks in particular, researchers have standardized the use of geometrically constrained patches, typically square, circular, or grid-shaped, that can be physically printed and affixed to target surfaces. To enhance the robustness of physical attacks, Sharif et al.[51] incorporated non-printability scores (NPS) into the loss calculations to better simulate real-world printing. Additionally, Expectation over Transformation (EOT)[52] has been used to mitigate the effects of environmental variations such as lighting, position, and angle, on adversarial attacks, and the Nested-AE algorithm[53] has been introduced to improve robustness across different distances and angles.

However, patch attacks exhibit notable limitations. In real-world applications, it is necessary to balance the content, shape, size, and placement of the patch in order to maximize adversarial effectiveness, inevitably making the patch conspicuous and easy to detect. In fact, studies on the interpretability of adversarial samples[54] indicate that many successful adversarial perturbations targeting object recognition tend to focus on object edges and texture regions. Perturbations affecting the shape characteristics of objects can have a substantial impact on the decision-making process of the classifier. Research on adversarial attacks leveraging generative models[55] further reveals that adversarial training implicitly captures and utilizes the shape and edge features of natural images, thereby enhancing both the realism and robustness of the attacks. This highlights an inherent trade-off between the effectiveness and stealthiness of patch-based attacks. For instance, effective attacks typically involve occluding the primary visual content of the sign in TSR, such as numbers, text, and patterns, which are generally key regions and commonly targeted for determining attack locations, as shown in Figure 1.

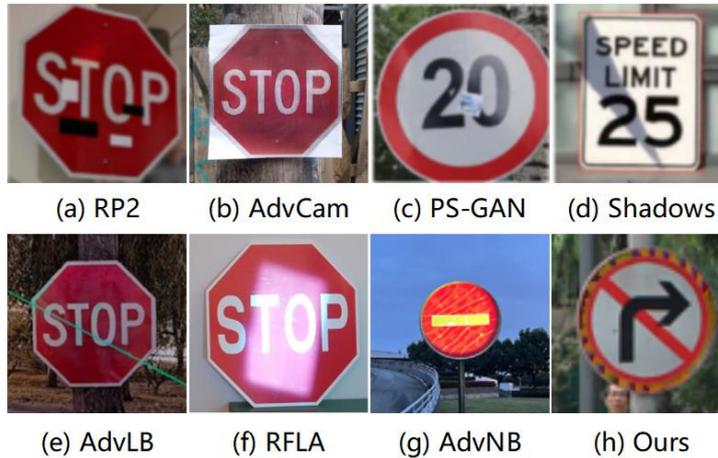

**Fig.1.** Visual comparison of physical attacks on TSR.

Research aimed at improving the stealthiness of traffic sign attacks has primarily focused on implementing attacks in more natural forms, such as using natural style transfer techniques like AdvCam[56], DAC[57], simulating lighting[58], or creating shadows[59] to maximize stealthiness. These attacks mimic the natural environment, such as rust, snow, lighting, and shadows; these elements may appear on traffic signs in real-world conditions. Optical-based methods are also a common physical deployment approach, including spoofing attacks using direct projections from projectors [15], and various lighting-based attacks, e.g., laser AdvLB[20], neon AdvNB[60], reflective light AdvCL[61], and RFLA[62]. These attacks induce misrecognition through optical control, leveraging their operational advantages of long-range applicability, temporary nature, and lack of residual trace to achieve superior concealment. Nevertheless, these attacks depend on projectors and specialized lighting equipment, which entail high deployment costs and impose strict requirements on the lighting conditions of the target environment. Furthermore, their effectiveness is significantly diminished when confronted with defense and detection mechanisms based on optical characteristics[63].

On the other hand, the potential for natural style in patch-based methods is significantly limited. Liu et al.[17] proposed a perceptually sensitive generative adversarial network (PS-GAN), through which the generated adversarial patches visually resemble common graffiti found on real-world signs. These naturally styled patches, achieved through advanced GAN architectures, can balance both naturalness and adversariality[64]. Additionally, instead of adding adversarial perturbations, there are methods based on real-world stickers, where the attack is implemented by iterating on the sticker's position and rotation angles[65]. Figure 1 displays visual examples of different physical adversarial attacks on traffic signs, including those generated by our proposed method.

Even though these patches maintain natural plausibility in their own images, they fail to achieve semantic consistency when placed on the natural background of traffic signs. Due to the fact that maximizing adversarial effectiveness often results in occluding the primary content of the sign, and given that adversarial patches require pre-deployment, sustaining their presence over extended periods without being detected remains a significant challenge. Furthermore, the limited adaptability of these patches' placement on traffic sign makes it difficult to execute large-scale, convenient, and effective attacks.

## 3. Methodology

This section provides a detailed description of the proposed traffic signs edge-mask patch attack method (TSEP-Attack). The general definition of adversarial patch attacks is first outlined, followed by an explanation of the attack paradigm based on edge masks. The algorithm for patch generation and the optimization of physical robustness are then presented. An overview of TSEP-Attack is illustrated in Figure 2.

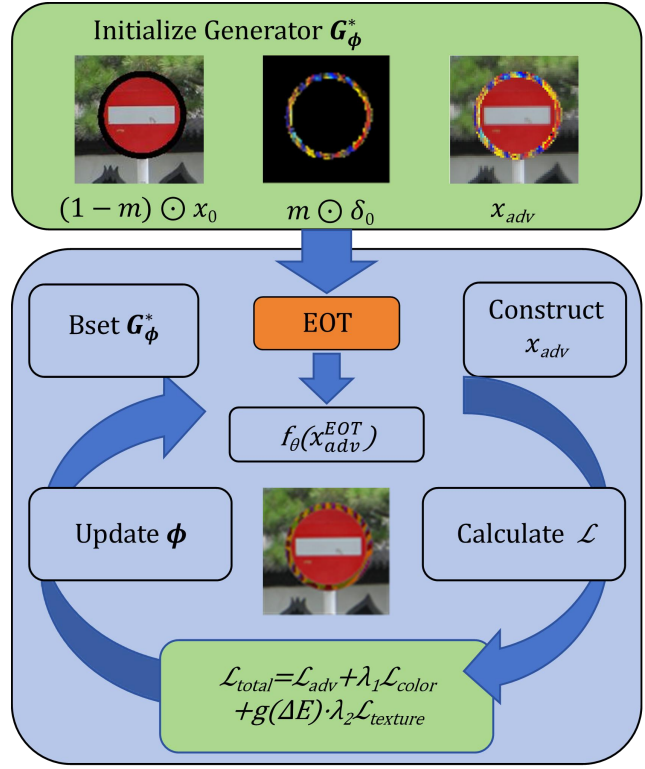

**Fig. 2.** An overview of TSEP-Attack.

*3.1. Adversarial patch attack for TSR*

TSR is implemented using image classification models. The DNN-based TSR model can be described as:

$$\hat{y} = f_\theta(x), \quad x \in X, y \in Y \tag{1}$$

where an image classifier: $f(\cdot):[0,1]^d \rightarrow R^N$ recognizes the input image $x \in R^{h \times w \times \varepsilon}$ as a class label $y_x \in (1,\ldots,N)$, where $h, w,$ and $\varepsilon$ represents the height, width, and channels of $x$, respectively, *and N* is the number of traffic sign categories. The classifier outputs the confidence score $f_r(x)$ for each class $r$, and assigns the class with the highest confidence score to the input image.

$$\hat{y} = \arg\max_{r=1,2,\ldots,N} f_r(x) \tag{2}$$

Adversarial patch attacks induce misclassification by modifying a small, contiguous region of the input images. Specifically, an adversarial patch can be represented as patch content, i.e., perturbation, $\delta \in [0,1]^d$ and a binary mask $m \in \{0,1\}^d$ representing the shape and position of the patch. The image after being disturbed by the adversarial patch can be formally defined as:

$$x_{adv} = m \odot \delta + (1-m) \odot x_0 \tag{3}$$

where $m$ represents the patch shape and position, $\delta$ represents the patch content, and $x_0$ is the original input image. Thus, the adversarial patch attack for an image $x_0$ with label y can be modeled as:

$$\min_{m,\delta} L\left(F(m \odot \delta + (1-m) \odot x_0), y'\right) \quad \text{s.t.} \ \|m\|_0 < \epsilon \tag{4}$$

where *L* is the adversarial loss, minimizing *L* will result in the misclassification of the victim model. $\|m\|_0 < \epsilon$ is an $l_0$-bounded constraint that controls the patch scale to preserve semantics; y' represents the true label $y_0$ in untargeted attacks, or the target class *t* in targeted attacks.

*3.2. Instance Segmentation-Based Edge Masks*

Distinguishing itself from general image classification, traffic sign classification involves categories that contain highly similar patterns, characterized by basic contour shapes — such as circles, triangles, rectangles, and octagons — combined with central emblematic figures. In practical physical deployments, traffic signs must adhere to stringent national standards. Variations within traffic sign image datasets primarily arise from sampling differences across diverse physical environments. This study confines the attack area to the peripheral region of the main traffic sign shape to enhance stealthiness. Moreover, contour-fitting adversarial patches simplify positioning compared to traditional patches and allow for large-scale deployment according to the standardized dimensions of traffic signs.

The Segment Anything Model (SAM) was utilized to acquire instance masks of traffic signs. SAM comprises an image encoder, a flexible prompt encoder, and a fast mask decoder. SAM produces high-quality object masks from input prompts such as points or boxes, and it can be used to generate masks for all objects in an image. It has been trained on the SA-1B dataset[22] , which includes 11 million images and 1.1 billion masks, and demonstrates strong zero-shot single-point performance across various segmentation tasks. Its zero-shot performance is highly robust and has been extensively validated on numerous new image distributions and tasks. In the dataset utilized, acquiring the region of interest (ROI) based on image annotations enables precise extraction of edge contours and instance masks for traffic signs.

The obtained instance masks were evaluated, and it was observed that they occasionally segmented internal elements such as text and patterns within the signs. Since the objective was to derive masks that exclusively fit the sign edges, direct processing of the instance masks could lead to the generation of multiple edges. Thus, by comparing contours identified through OpenCV, only the largest outer contour was retained, and the instance mask was filled to form a solid region. Based on this, an annular patch mask width is determined as a percentage of the input image dimensions. Through inward erosion and outward expansion operations, an annular mask that precisely fits the traffic sign's boundary is generated.

*3.3. Adversarial patch sample generation network*

In generative adversarial attacks, generative networks are commonly used to enhance stealthiness. However, GANs—which consist of both a generator and a discriminator—often suffer from slow convergence due to optimization difficulties in reaching Nash equilibrium. While the generator aims to mislead DNN classifiers, the discriminator works to minimize discrepancies between generated and real data. This adversarial dynamic frequently leads to gradient conflicts and training instability. In our setting, patch location and shape are rigidly constrained to avoid occluding key regions of the traffic sign. As a result, GANs struggle to balance stealth and effectiveness under such limitations, especially compared to methods allowing flexible optimization of position and shape.

Therefore, this work employs a U-Net generator enhanced with VGG natural texture priors to produce adversarial patches. A multi-level loss function incorporating color, texture, and frequency domain constraints is applied to ensure visual stealth and physical robustness.

3.3.1. Generator network architecture

This work adopts a texture-aware adversarial patch generator built upon a U-Net tailored to 128×128 inputs. The network accepts an 11-channel tensor composed of latent noise, a scalar mask, a data-driven texture prior, and the masked original image, thus grounding synthesis in both stochastic and image-specific cues. The encoder comprises three double-convolution blocks interleaved with 2×2 max pooling, shrinking the spatial scale from 128×128 to 16×16 while expanding channels from 32 to 256. A bottleneck double-convolution at the coarsest scale aggregates global semantics.

The decoder mirrors this structure: each of the three stages begins with a 2×2 transposed convolution to upsample, concatenates the result with the encoder feature of the same resolution, and applies another double-convolution block. This preserves high-frequency texture cues while propagating bottleneck information. A final 3×3 convolution followed by Tanh outputs a three-channel residual, which is scaled and added to the blended texture prior before being masked and clipped to [−1, 1].

Texture priors are constructed by sampling multiple patches within the ring mask from the original image, resizing, and averaging them; the prior is then linearly blended with the masked original image using prior_blend. Both latent noise and priors are bilinearly resized to the working resolution to ensure alignment. Overall, the architecture retains U-Net's symmetric encoder–decoder advantages while injecting mask guidance and learned priors to generate high-fidelity adversarial patches with manageable computational cost.

The network input is designed for multimodal fusion:

$$\mathbf{x}_{in} = [\mathbf{z}, \mathbf{m}, \mathbf{p}_{tex}, \mathbf{x}] \in \mathbb{R}^{11 \times H \times W} \tag{5}$$

Each modality serves a distinct functional role:

The latent variable $\mathbf{z} \in \mathbb{R}^{4 \times H \times W}$ is sampled from an isotropic Gaussian distribution $\mathbf{z} \sim \mathcal{N}(0, \mathbf{I})$, introducing stochasticity to enable one-to-many mapping. This allows the generator to produce diverse adversarial patches from the same input image, thereby avoiding mode collapse.

The mask $\mathbf{m} \in [0,1]^{1 \times H \times W}$, extracted using the SAM model, precisely specifies the spatial region where the patch is applied. Compared to random rectangular patches, this instance-aware mask better conforms to the target object's boundaries, reduces interference from background areas, and improves boundary blending.

The texture prior $\mathbf{p}_{tex} \in \mathbb{R}^{3 \times H \times W}$ guides style consistency, while the original RGB image $\mathbf{x} \in \mathbb{R}^{3 \times H \times W}$ provides global contextual information. This enables the generator to account for overall lighting conditions, color distribution, and scene semantics, thereby producing adversarial content that is visually coherent with the scene.

The output adopts a residual learning strategy, predicting only the correction to the texture prior. This approach reduces the learning burden while ensuring visual stealth. In terms of technical implementations, Group Normalization replaces Batch Normalization to accommodate the stability requirements of small-batch training, and the GELU activation function provides smoother gradient propagation compared to ReLU, which is conducive to texture detail modeling.

3.3.2. Loss Function

The core challenge in adversarial attacks lies in balancing the conflicting objectives of attack effectiveness and visual stealth: over-optimizing the attack loss may produce noticeably perceptible adversarial noise, while over-constraining stealth may fail to deceive the target model. To address this, a multi-level composite loss function was designed, achieving an optimal balance through hierarchical constraints and adaptive scheduling:

$$\mathcal{L}_{total} = \mathcal{L}_{adv} + \lambda_1 \mathcal{L}_{color} + g(\Delta E) \cdot \lambda_2 \mathcal{L}_{texture} \tag{6}$$

where $\mathcal{L}_{adv}$ denotes the adversarial loss ensuring model misclassification; $\mathcal{L}_{color}$ represents the color constraint loss maintaining overall color matching;

$\mathcal{L}_{\text{texture}}$ indicates the texture constraint loss optimizing fine-grained visual consistency; g(ΔE) serves as an adaptive gating function based on color difference; and $\lambda_1, \lambda_2$ are weighting hyperparameters.

This design follows a hierarchical optimization principle: Color constraint prioritizes macroscopic color coordination, while texture constraint further refines microscopic texture details once color proximity is achieved. The gating function $g(\cdot)$ enables dynamic weight adjustment, adaptively allocating constraint intensity according to the current optimization state to avoid the limitations of fixed weights, such as initial training difficulties or insufficient stealth in later stages.

1. Adversarial Loss

Given a target classifier $f_\theta: \mathbb{R}^{3 \times H \times W} \to \mathbb{R}^C$, an adversarial image $\mathbf{x}_{adv} = (1 - \mathbf{m}) \odot \mathbf{x} + \mathbf{p}$, and the true class $y$, our attack objective is to minimize the model's confidence in the true class. Adversarial examples in the digital domain face multiple challenges when deployed in the physical world: the printing-capture process introduces color distortion, camera viewpoint changes cause geometric deformations, and environmental lighting affects image brightness and contrast. These factors significantly reduce the attack success rate. To address this, Expectation Over Transformation (EOT) is adopted, applying random transformations to adversarial samples during training. This enables the generator to learn how to produce patches robust to various transformations, thereby enhancing physical-world robustness.

$$\mathcal{L}_{adv} = \mathbb{E}_{\mathcal{T} \sim \tau}[f_\theta(\mathcal{T}(\mathbf{x}_{adv}))_y] \quad (7)$$

Here, $\mathcal{T}$ denotes a random transformation function, and $\tau$ represents the transformation distribution (including rotation, scaling, and brightness adjustment). $f_\theta(\cdot)_y$ indicates the logit output value for the target class $y$. Compared to traditional cross-entropy loss $-\log P(y|\mathbf{x}_{adv})$ or targeted attack loss $-\log P(y_{target}|\mathbf{x}_{adv})$, directly minimizing the logit value of the true class is more concise and efficient. It eliminates the need for softmax normalization and logarithmic operations, leading to more stable gradient computation. Adopting a non-targeted attack also provides greater freedom for optimization.

In practice, we approximate the expectation by sampling $N$ transformations $\{\mathcal{T}_i\}_{i=1}^N$.

$$\mathcal{L}_{adv} \approx \frac{1}{N} \sum_{i=1}^N f_\theta(\mathcal{T}_i(\mathbf{x}_{adv}))_y \quad (8)$$

Each transformation $\mathcal{T}_i$ is independently sampled with rotation angle $\theta_i \sim \mathcal{U}(-15°, 15°)$, scaling factor $s_i \sim \mathcal{U}(0.9, 1.1)$, brightness factor $b_i \sim \mathcal{U}(0.8, 1.2)$, and contrast factor $c_i \sim \mathcal{U}(0.9, 1.1)$. The choice of $N = 4$ is based on a trade-off between efficiency and effectiveness: a larger $N$ improves robustness but linearly increases computational cost.

During backpropagation, gradients are averaged across multiple transformation samples. This steers the optimization direction toward solutions that are robust to various physical transformations, thereby enhancing the success rate of real-world deployment.

$$\nabla_\mathbf{p} \mathcal{L}_{adv} = \frac{1}{N} \sum_{i=1}^N \nabla_\mathbf{p} f_\theta(\mathcal{T}_i(\mathbf{x}_{adv}))_y \quad (9)$$

2. Color Constraint Loss

Color is the primary factor affecting visual stealth, as noticeable differences between patch colors and the background can immediately attract attention. A three-tiered progressive color constraint is designed: numerical range constraint, perceptual color difference constraint, and saturation constraint, each ensuring color coordination from different perspectives.

1) RGB Range Constraint

This limits pixel values within a reasonable range to avoid generating oversaturated or extreme colors:

$$\mathcal{L}_{rgb} = \frac{1}{|\mathcal{M}|} \sum_{(i,j) \in \mathcal{M}} [\max(0, |\mathbf{p}_{ij}| - b_{rgb})] \quad (10)$$

where $\mathcal{M}$ represents the set of pixels in the patch region, $\mathbf{p}_{ij} \in [-1,1]^3$ denotes the RGB pixel value at position $(i,j)$, and $b_{rgb}$ is the soft boundary threshold. The ReLU function $\max(0, \cdot)$ implements a soft constraint, which provides continuous gradients compared to hard clipping $clip(\mathbf{p}, -b, b)$, thereby avoiding optimization stagnation at boundaries.

2) LAB Color Space Constraint

Euclidean distance in RGB space does not align with human perceptual characteristics. For example, the perceived difference between (255,0,0) and (200,0,0) is much smaller than that between (128,0,0) and (128,128,0), yet the Euclidean distance in RGB suggests the opposite. The CIE LAB color space is designed to be perceptually uniform, meaning the Euclidean distance between colors—known as $\Delta E$—corresponds linearly to the difference perceived by the human eye[66].

The RGB values of the patch and background regions are first converted to the LAB space via gamma correction, intermediate XYZ space transformation, and nonlinear mapping. The weighted average LAB values of the patch region and the background annular region are then computed as follows:

$$\bar{\mathbf{L}}_{patch} = \frac{\sum_{(i,j) \in \mathcal{M}} \text{RGB2LAB}(\mathbf{p}_{ij}) \cdot \mathbf{m}_{ij}}{\sum_{(i,j) \in \mathcal{M}} \mathbf{m}_{ij}} \quad (11)$$

$$\bar{\mathbf{L}}_{bg} = \frac{\sum_{(i,j) \in \mathcal{M}_{ring}} \text{RGB2LAB}(\mathbf{x}_{ij}) \cdot \mathbf{m}_{ij}^{ring}}{\sum_{(i,j) \in \mathcal{M}_{ring}} \mathbf{m}_{ij}^{ring}} \quad (12)$$

where $\mathcal{M}_{ring}$ denotes the annular background region. The color difference $\Delta E$ is defined as the Euclidean distance in the LAB space:

$$\Delta E = \|\bar{\mathbf{L}}_{patch} - \bar{\mathbf{L}}_{bg}\|_2 = \sqrt{(\Delta L)^2 + (\Delta a)^2 + (\Delta b)^2} \quad (13)$$

The loss function is designed as a penalty for exceeding a threshold:

$$\mathcal{L}_{lab} = \max(0, \Delta E - \tau_{lab}) \tag{14}$$

The threshold $\tau_{lab}$ is set based on the CIE color difference perception standard [67].

3) Saturation Constraint

Excessively high saturation can make patches appear overly vivid and difficult to blend into real backgrounds. Pixel saturation is measured by the span of RGB channels:

$$S(\mathbf{p}_{ij}) = \max_c(\mathbf{p}_{ij}) - \min_c(\mathbf{p}_{ij}) \tag{15}$$

Since the generator output is linearly mapped to the range [0,1], the difference between the maximum and minimum values directly describes saturation without additional normalization.

$$\mathcal{L}_{sat} = \max(0, \bar{S} - \tau_{sat}), \tag{16}$$

The regional average saturation is normalized using mask weights:

$$\bar{S} = \frac{\sum_{(i,j)\in\mathcal{M}} S(\mathbf{p}_{ij}) \cdot \mathbf{m}_{ij}}{\sum_{(i,j)\in\mathcal{M}} \mathbf{m}_{ij}} \tag{17}$$

The threshold is derived from statistics of background regions in the TSRD traffic sign dataset.

The weights for the three color constraints are set as $\lambda_{rgb} = 0.05, \lambda_{lab} = 0.10, \lambda_{sat} = 0.03$, with a weight ratio of approximately 1:2:0.6. The LAB constraint has the highest weight as it directly corresponds to human perception; the RGB constraint follows as a numerical safeguard; and the saturation constraint has the lowest weight, serving an auxiliary role.

3. Texture Constraint Loss

Building upon overall color matching, consistency in texture details further determines the visual stealth of the patch. Three complementary texture constraints are designed to enforce consistency between the generated texture and the background from three perspectives: style statistics, frequency domain characteristics, and spatial smoothness.

1) Gram Matrix Style Loss

The Gram matrix of intermediate feature layers from a pre-trained VGG19 network is used to represent texture style. The Gram matrix captures correlations between feature channels, effectively characterizing the statistical properties of textures while ignoring specific spatial arrangements.

Given a feature extractor $\phi: \mathbb{R}^{3 \times H \times W} \to \mathbb{R}^{D \times H' \times W'}$, the Gram matrix is defined as the inner product of the feature maps:

$$\mathbf{G}(\mathbf{F}) = \frac{1}{CHW}\mathbf{F}\mathbf{F}^\top \in \mathbb{R}^{D \times D} \tag{18}$$

where $\mathbf{F} = reshape(\phi(\mathbf{x}), [D, H'W']) \in \mathbb{R}^{D \times H'W'}$ is the reshaped feature matrix, and the normalization factor $CHW$ ensures scale invariance. The $(i,j)$-th element $G_{ij}$ of the Gram matrix represents the correlation between the $i$-th and $j$-th feature channels:

$$G_{ij} = \frac{1}{CHW}\sum_{k=1}^{H'W'} F_{ik} F_{jk} \tag{19}$$

The diagonal elements $G_{ii}$ reflect the energy of the $i$-th channel, while the off-diagonal elements $G_{ij}$ represent the covariance between channels. This second-order statistic effectively captures repetitive patterns and structural characteristics of textures, independent of specific spatial locations, making it suitable for texture matching tasks.

The style loss is defined as the Frobenius norm distance between the Gram matrices of the patch region and the background region:

$$\mathcal{L}_{gram} = \|\mathbf{G}(\phi(\mathbf{p}_{patch})) - \mathbf{G}(\phi(\mathbf{x}_{bg}))\|_F^2 \tag{20}$$

where $\mathbf{p}_{patch} = \mathbf{m}_{ring} \odot \mathbf{p}$ denotes the masked patch region, $\mathbf{x}_{bg} = \mathbf{m}_{ring} \odot \mathbf{x}$ represents the background annular region, and $\|\cdot\|_F$ is the Frobenius norm.

2) FFT Frequency Domain Loss

While the Gram matrix primarily captures spatial correlations, a frequency domain constraint is incorporated to characterize the periodicity and frequency distribution of textures. The magnitude spectrum $|\mathcal{F}(\mathbf{x})|$ reflects the intensity of each frequency component:

$$\mathcal{L}_{fft}^{mse} = \|\mathbf{A}_{patch} - \mathbf{A}_{bg}\|_2^2 \tag{21}$$

low frequencies correspond to smooth regions and overall brightness, while high frequencies correspond to edges and texture details. The frequency domain distance is defined as the L2 distance of the magnitude spectra:

$$\mathbf{A} = |\mathcal{F}(\mathbf{x})| \in \mathbb{R}^{3 \times H \times W} \tag{22}$$

where the 2D FFT is computed independently for each RGB channel. This loss encourages the patch and background to exhibit similar spectral distributions. A boundary penalty term is added:

$$\mathcal{L}_{fft}^{margin} = \|\max(0, \mathbf{A}_{patch} - \mathbf{A}_{bg} - \epsilon_{fft})\|_1 \tag{23}$$

where $\epsilon_{fft} = 0.05$ is the tolerance margin. This term penalizes components of the patch magnitude spectrum that exceed the background spectrum plus the margin, allowing slight enhancement while preventing excessive amplification. The complete FFT loss is:

$$\mathcal{L}_{fft} = \mathcal{L}_{fft}^{mse} + \lambda_{margin}\mathcal{L}_{fft}^{margin} \tag{24}$$

With $\lambda_{margin} = 1.0$, the two terms contribute comparably.

3) Total Variation Regularization

Total Variation (TV) regularization encourages spatial smoothness in the generated patch, suppressing high-frequency noise and checkerboard artifacts. The discrete TV is defined as the sum of the L1 norms of differences between adjacent pixels:

$$\mathcal{L}_{tv} = \frac{1}{|\mathcal{M}|} \sum_{(i,j)\in\mathcal{M}} \left(|\mathbf{p}_{i+1,j} - \mathbf{p}_{i,j}| + |\mathbf{p}_{i,j+1} - \mathbf{p}_{i,j}|\right) \tag{25}$$

where $|\cdot|$ denotes the element-wise absolute value. This formulation computes first-order gradients in horizontal and vertical directions, with the L1 norm (compared to L2) better preserving edges.

4. Adaptive Scheduling Mechanism

Fixed-weight loss functions face a dilemma: if texture constraints are too strong initially, optimization becomes difficult, hindering the reduction of the attack loss; if texture constraints are too weak later, stealth is compromised. A two-layer adaptive mechanism is designed: a color difference-based gating function (runtime adaptation) and a training step-based warmup scheduler (training-phase adaptation). This achieves a dynamic balance, shifting priority from attack effectiveness to joint optimization of attack and stealth.

1) Color Difference-Based Gating Function

Texture constraints are meaningful only when colors are already close to the background; applying them too early can interfere with color optimization. An adaptive gating function dynamically adjusts the weight of the texture loss based on the current color difference $\Delta E$:

$$g(\Delta E) = \sigma(k(\tau_{lab} - \Delta E)) = \frac{1}{1 + \exp(-k(\tau_{lab} - \Delta E))} \tag{26}$$

where $\sigma(\cdot)$ is the sigmoid function, $k$ is the steepness parameter, and $\tau_{lab}$ is the color difference threshold.

When $\Delta E \gg \tau_{lab}$ (large color difference): $g(\Delta E) \to 0$, the texture constraint weight approaches zero, prioritizing color matching. When $\Delta E \approx \tau_{lab}$ (color near the threshold): ($g(\Delta E) \approx 0.5$), the texture constraint begins to take effect. When $\Delta E \ll \tau_{lab}$ (color matched): $g(\Delta E) \to 1$, the texture constraint weight reaches its maximum, finely optimizing texture details.

The gating mechanism simulates human visual attention: when the color difference $\Delta E$ between two regions is significant, texture details are ignored; only when colors are similar do texture differences become noticeable. This hierarchical optimization strategy aligns with the processing flow of the human visual system and follows the "coarse-to-fine" principle in adversarial example design.

2) Warmup-Based Training Scheduler

Even with the gating function, gradients in early training may still be dominated by texture constraints due to their high dimensionality and large values. A linear warmup strategy is applied to texture-related loss terms:

$$\lambda_{gram}^{eff}(t) = \lambda_{gram} \cdot \alpha_{gram}(t) \tag{27}$$

$$\alpha_{gram}(t) = \min\left(1, \frac{t}{T_{gram}}\right) \tag{28}$$

where $t$ is the current training step, $T_{gram}$ is the total warmup steps, and $\alpha_{gram}(t)$ is the warmup coefficient. Similarly, warmup is applied to the FFT loss and TV regularization, while the color constraint and attack loss take full effect from the first step.

Additionally, a warmup strategy is applied to the generator's residual scaling factor:

$$\alpha_{res}(t) = \min\left(1, \frac{t}{T_{res}}\right) \tag{29}$$

where $\alpha_{res}(t)$ is the warmup coefficient and $T_{res}$ is the total warmup steps. This makes the early training rely primarily on the texture prior, gradually enhancing the generator's modification capability later.

3.3.3. TSEP-Attack Algorithm

The TSEP-Attack algorithm operates in batch mode and coordinates a conditional U-Net generator through nested epoch, batch, and sample loops. It integrates EOT for physical robustness and multi-layer perceptual and color constraints to generate adversarial patches suitable for real-world deployment. The pseudocode outlines the input and output specifications, nine procedural stages, and a checkpointing mechanism, forming a structured pipeline: generate, transform, regularize, update, and save.

The algorithm steps are as follows:

| Algorithm 1: TSEP-Attack algorithm |
|---|
| Input: target model $f_\theta$, dataset $D = (x, m, y)$, max epochs $T_{max}$, hyperparameters $\mathcal{C}$ |
| Output: trained generator $G_\phi^*$ |
| 1: Initialize $G_\phi$, EOT module $E$, VGG subnetwork $\Phi$, optimizer Adam, warmup steps $T_{warm}$ |
| 2: for epoch = $(1 \ldots T_{max})$ do |
| 3:     for batch $(B \subset D)$ do |

```
4:      for ((x, m, y) ∈ B) do
5:          global_step( ← )global_step( + 1)
6:          m_seg ← LoadSegMask(x); m_ring ← ExtractRing(m_seg)
7:          p_tex ← BuildTexturePrior(x, m_ring)
8:          z~N(0, I); α_res ← Warmup(global_step, T_warm^res)
9:          δ ← clip(G_Φ(z, m, p_tex, x; α_res), −1,1)
10:         {x_adv^(i)}_{i=1}^{N_eot} ← E(x, δ, m)
11:         L_adv ← (1/N_eot) Σ_{i=1}^{N_eot} f_θ(x_adv^(i))_y
12:         Compute(L_rgb, L_lab, L_sat)(color constraints)
13:         α_gram, α_fft ← Warmup(global_step, T_warm^{gram/fft})
14:         L_gram, L_fft, L_tv ← Texture Losses(Φ, δ, x, m_seg, m_ring)
15:         g ← σ(k · (τ_lab − ΔE(δ, x)))
16:         L_total ← L_adv + λ_rgb L_rgb + λ_lab L_lab + λ_sat L_sat )
17:             + g(α_gram λ_gram L_gram + α_fft λ_fft L_fft + α_gram λ_tv L_tv)
18:         Backpropagate L_total, clip gradients, update ϕ
19:     end for
20:     end for
21:     if best performance then Save G_ϕ, optimizer
22: end for
23: return G_ϕ*
```

1. Initialization: All components are initialized before training begins: a U-Net generator with a latent dimension of 4 and 32 base channels is instantiated, an EOT module is enabled, and the first nine layers of VGG19 are used to extract texture features. The Adam optimizer is configured with a learning rate of $5×10^{-4}$ for generator parameters. Global step and warmup counters are defined for Gram matrix, FFT, and residual components, set to 600, 600, and 200 steps respectively, ensuring gradual incorporation of residual injection and texture/frequency regularization.

2. Sampling and Training：For each sample, the algorithm follows these steps:1). Loads the instance segmentation mask, extracts a ring-shaped mask, and constructs a texture prior; 2). Samples latent noise from a standard Gaussian distribution, scales residuals, and generates a clipped adversarial patch; 3). Passes the patch through the EOT module to produce multiple transformed adversarial views; the attack loss is defined as the average logits over these views; 4). Accumulates RGB, LAB, and saturation constraints, along with Gram matrix, FFT, and total variation regularizers. A color-difference gating mechanism and warmup schedulers modulate these terms to form the total loss.

Each step concludes with numerical stability checks, gradient backpropagation, gradient norm clipping, parameter updates, and periodic checkpoint saving.

3. Complexity Analyses: Initialization happens once per run, whereas the dominant cost stems from $T_{max}$ iterative updates: each iteration requires one U-Net forward and one classifier inference, amplified by $N_{eot}$, while texture penalties add truncated VGG19, FFT, and TV computations. Denoting generator and classifier costs by $C_G$ and $C_{f_θ}$, the overall time complexity is $O(T_{max} · |D| · (N_{eot}(C_G + C_{f_θ}) + C_{loss}))$; spatial complexity is dictated by network activations/gradients, i.e., $O(θ)$ dominated by executing the target model.

## 4. Experiments

This section first presents the experimental setup and parameter settings, then demonstrates the validation and comparative experiments along with analysis of our attack method, and finally conducts the ablation study.

*4.1. Experimental setup*

1. Datasets: The benchmark datasets used in this study are TSRD, which contains common prohibition signs, warning signs, and indication signs, for a total of 58 categories of traffic sign images. The training set contains 4,170 samples, and the validation set contains 1,994 samples. To ensure equitable evaluation of adversarial attack performance given the imbalanced class distribution in the dataset, a standardized sampling approach was applied to construct the test set. From both the training and validation sets, five correctly classified images per class were selected across all victim recognition networks. Categories with insufficient numbers of images were excluded, resulting in a final adversarial test set comprising 265 traffic sign images spanning 53 distinct classes.

2. Models: A diverse set of victim recognition networks was utilized for evaluation. Given the inconsistent image sizes in the datasets, all input images were normalized to $128×128$ pixels before being fed into the models. Four recognition models were constructed for the experiments, based on GTSRB-CNN, ResNet18, ResNet50 [3], and MobileNetV3 [68] architectures. To enhance the robustness of the recognition models, the training set was augmented through rotation, cropping, translation, brightness adjustment, and contrast adjustment. After removing invalid samples, a total of 10,008 samples (including augmented data) were retained for pre-training the victim classification models, with 833 samples allocated for validation and 1,993 samples for testing. As demonstrated in Table 1, all four models achieved excellent recognition accuracy on the test datasets.

3. Comparison Methods:A comparative analysis was conducted to evaluate the attack capabilities of PGD, Google AP, and ShadowAttack on our adversarial test set. To ensure fairness in comparison, all methods were trained using identical epochs and batch sizes, maintaining a uniform query budget during attack optimization. Performance was subsequently validated on the standardized adversarial test set.

4. Evaluation Metrics：To evaluate and compare the attack effectiveness of different methods, the Attack Success Rate (ASR) was adopted, calculated as the ratio of successfully attacked images to the total number of attacked images within a specified number of epochs. However, a single evaluation metric is insufficient to fully characterize the complexity of adversarial examples under human visual perception. Therefore, a comprehensive quantitative analysis was conducted using multiple metrics:The Structural Similarity Index (SSIM)[69] was employed to assess the preservation of structural information, simulating human visual sensitivity to structural distortions. The Feature Similarity Index (FSIM)[70] was utilized to measure the similarity of low-level features (such as edges and contours), reflecting the primary cognitive processes of the human visual system. The Gradient Magnitude Similarity Deviation (GMSD)[71] was introduced to quantify deviations in gradient structure. The Mean Absolute Error (MAE) was computed to provide a baseline for pixel-level absolute differences. This multi-faceted evaluation framework enables a more holistic assessment of both the attack potency and perceptual stealth of the generated adversarial examples.

5. Physical attack test: The adversarial samples generated by TSEP-Attack were physically printed on A4 paper and captured using a mobile phone from various distances and viewing angles. The captured images were subsequently cropped before being fed into the models for validation. All images were acquired using a Xiaomi 14 smartphone at a resolution of 3072×3072 pixels.

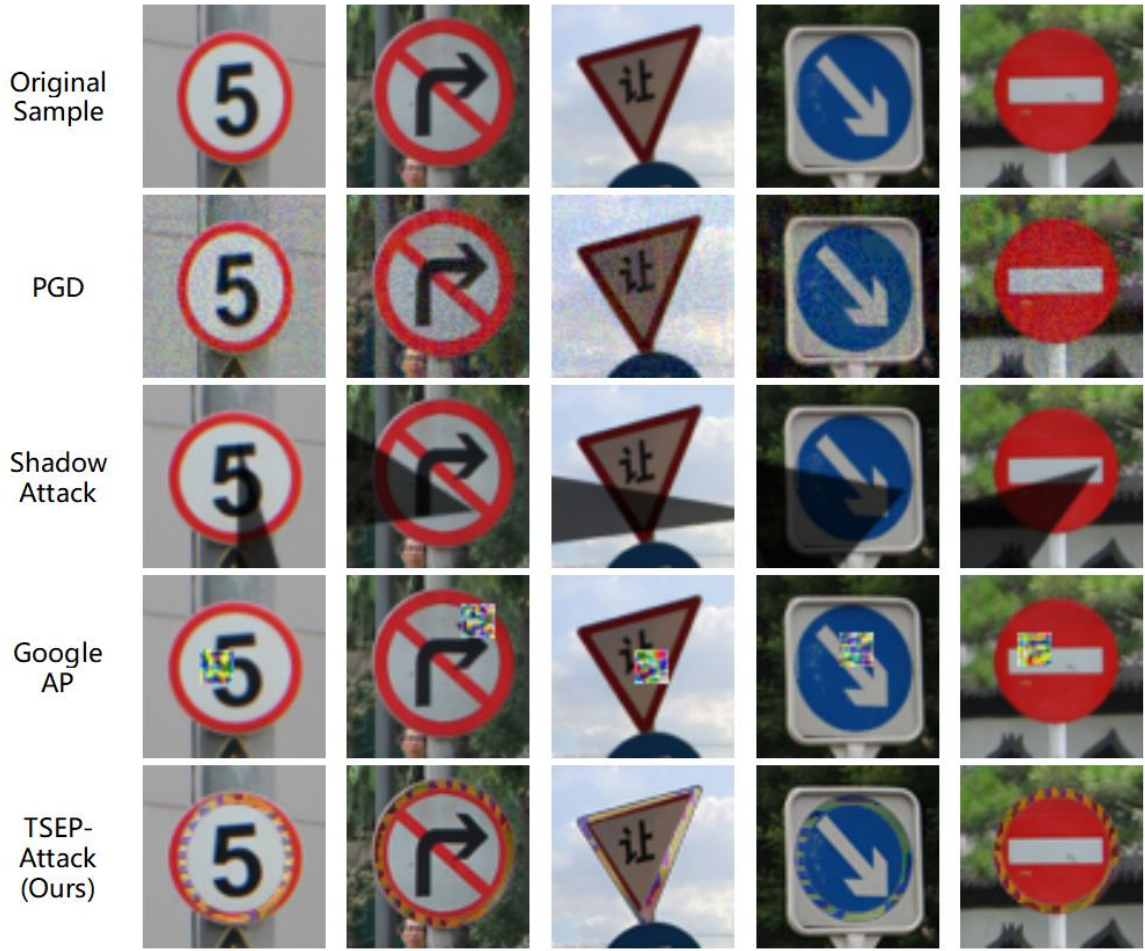

**Fig.3.** Visual comparison of the four attack methods.

**Table 1 - Performance of Pretrained Models**

| Models | Gtsrb-CNN | MobileNetV3 | ResNet18 | ResNet50 |
|---|---|---|---|---|
| Accuracy(%) | 98.39 | 96.84 | 99.40 | 98.09 |

### 4.2.Parameter settings

The compared methods include white-box attacks (PGD, Google AP) and the black-box attack ShadowAttack. Due to methodological differences, reasonable modifications were made to the open-source codes to adapt them to our data processing pipeline, while unifying the computation of attack iterations. Parameters followed recommended values from the respective papers unless otherwise specified. Specifically, the GAP patch size was set to 24×24 for the CTSRD dataset, and the perturbation threshold in PGD attacks was set to ε=0.1.

In TSEP-Attack, the annular edge mask width ratio $r$ was set to 0.04, defined as a percentage of the shorter side length of the input image. Due to variations in edge shapes, the proportion of mask pixels fluctuates, with an average value below 10%. It was observed that the edge mask width ratio significantly influences the attack success rate—adopting a larger ratio generally improves attack effectiveness. The current parameter was determined by balancing edge fitting accuracy and visual stealth through empirical evaluation on sample images.

The parameter value of $b_{rgb} = 0.75$ for the RGB range constraint was selected based on two considerations. First, it avoids extreme pixel values that are susceptible to inaccurate color reproduction in the printing process. Second, the resulting interval [32, 223] preserves approximately 75% of the full RGB dynamic range, which prevents extremes while maintaining sufficient color expressiveness.

The threshold $\tau_{lab} = 12.0$ for the LAB color space constraint was selected based on the CIE color difference perception standard [sharma2005ciede2000]: $\Delta E < 1$: Imperceptible to the human eye; $1 \leq \Delta E < 2$: Perceptible only by experts under strict conditions; $2 \leq \Delta E < 10$: Slight difference, difficult for non-experts to describe; $10 \leq \Delta E < 49$: Clearly noticeable and easily distinguishable; $\Delta E \geq 49$: Completely different colors. Setting $\tau_{lab} = 12.0$ is slightly more lenient than the "clearly noticeable" threshold to balance stealth and attack effectiveness: an overly strict threshold would reduce the ASR, while an overly loose threshold would increase human detection rates.

In the color difference-based gating function, the parameter $k = 4.0$ controls the steepness of the sigmoid: A larger $k$ makes $g(\Delta E)$ change more abruptly near $\Delta E = \tau_{lab}$, resembling a hard switch. A smalle r$k$ results in a smoother transition. Choosing $k = 4.0$ allows the gating function to complete a smooth transition from 0 to 1 within the range $\Delta E \in [\tau_{lab} - 2, \tau_{lab} + 2]$ (approximately $\Delta E \in [10,14]$). This avoids training instability caused by hard switching while ensuring sufficiently fast response.

In the Warmup-Based Training Scheduler, the total warmup steps for each phase were set to specific intervals: $T_{gram} = 600$, $T_{res} = 200$, $T_{fft} = 600$, $T_{tv} = 600$. By integrating the gating mechanism with warmup scheduling, the training process is naturally divided into three distinct phases:

Phase 1 (0-200 steps): $\alpha_{res} < 1, \alpha_{gram} < 0.33, g \approx 0$: Primarily relies on the texture prior, optimizing the attack loss and color constraint.

Phase 2 (200-600 steps): $\alpha_{res} = 1$, $\alpha_{gram} \in [0.33,1]$ in [0.33, 1], $g$ gradually increases: The generator modifies at full capacity, and texture constraints are progressively strengthened.

Phase 3 (600+ steps): $\alpha_{res} = 1, \alpha_{gram} = 1, g \approx 1$: All constraints are fully active, jointly optimizing attack effectiveness and stealth.

*4.3. Experimental results*

4.3.1. Attack performance and comparison

Figure 4 demonstrates the attack success rate performance of our method against different traffic sign recognition models on the adversarial test set constructed based on TSRD, showing its progression across iteration rounds. Table 2 presents the achievable range of attack success rates during 100 training epochs.

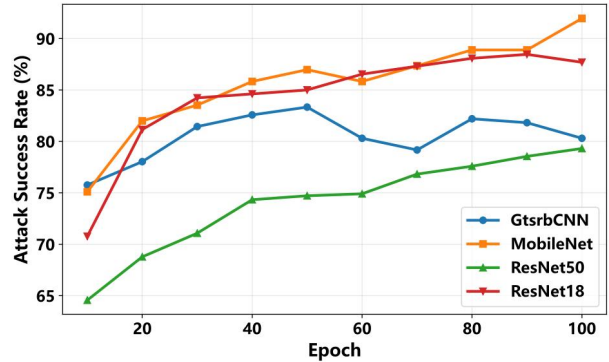

**Fig. 4**. Performance of digital attack on different models

**Table 2 - ASR(%) on different models**

| Models | Min | Max | Range |
| --- | --- | --- | --- |
| GtsrbCNN | 31.44 | 84.09 | 52.65 |
| MobileNet | 52.11 | 91.95 | 39.85 |
| ResNet50 | 44.06 | 80.08 | 36.026 |
| ResNet18 | 40.38 | 89.23 | 48.85 |

The results indicate that our attack method achieves over 80% success rate against various traffic sign recognition models in the digital domain within certain iteration rounds, reaching 89.23% and 91.95% on MobileNet and ResNet18 respectively, demonstrating strong white-box attack capability across multiple models. Furthermore, our method attains considerable attack success rates even during initial iterations and exhibits rapid growth in early stages. Through mid-to-late phases, it maintains stable improvement guided by our composite loss function, with some instances reaching convergence. For ResNet50 and MobileNet networks, the attack success curves still show upward trends after 100 training epochs, indicating promising attack potential.

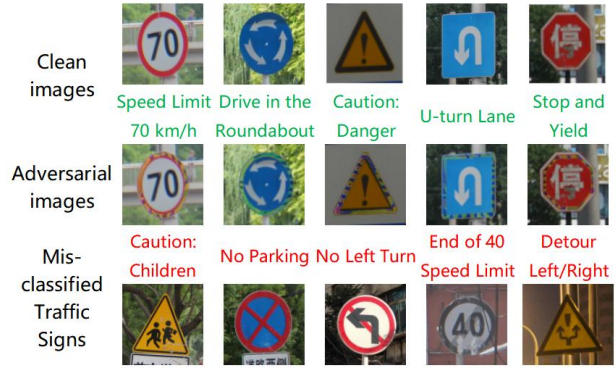

**Fig.5** Adversarial examples and the mis-classified traffic signs.

Furthermore, we compared our method with other attack approaches on the ResNet18 model. Figure 6 presents the attack success rates of different methods on the adversarial test set. Our TSEP-Attack demonstrates comparable attack capability to PGD and ShadowAttack, all outperforming Google AP. It should be noted that among the compared methods, PGD is a sparse global digital attack, while TSEP-Attack, ShadowAttack, and Google AP all incorporate EOT transformations to simulate physical-world conditions, indicating our method's strong potential for physical implementation. Figure 3 displays adversarial examples generated by the four attack methods alongside their original images.

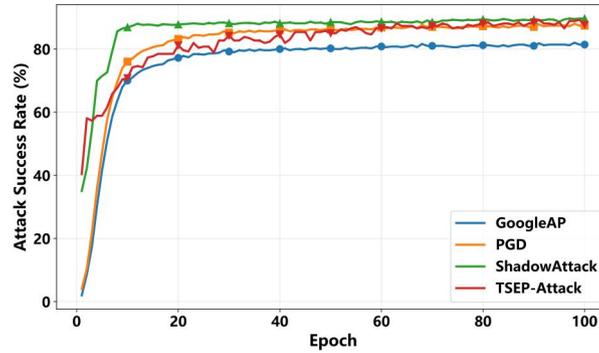

**Fig. 6.** ASR comparison for four methods.

To quantitatively assess the visual stealthiness of adversarial examples, this study employs a comprehensive analysis based on four perceptual quality metrics: higher values of Structural Similarity (SSIM) and Feature Similarity (FSIM) indicate closer resemblance to the original image and enhanced stealthiness, whereas lower values of Gradient Magnitude Similarity Deviation (GMSD) and Mean Absolute Error (MAE) reflect reduced image distortion and superior concealment characteristics.

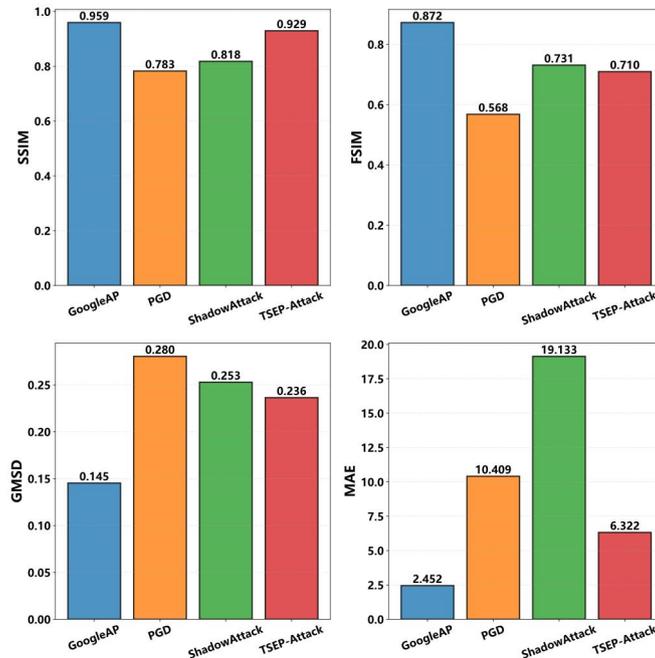

**Fig. 7.** Stealthiness comparison for three methods.

Figure 7 presents the metric comparisons of the four methods to evaluate their stealthiness performance.Based on the mean value analysis, TSEP-Attack demonstrates excellent and consistent stealthiness across all four metrics. Specifically, in terms of structural preservation, it achieves an SSIM mean of 0.929, which is second only to GoogleAP (0.959) and significantly higher than PGD (0.783) and ShadowAttack (0.818). This indicates its ability to effectively maintain the authenticity of the global image structure, exhibiting strong structural deception. At the feature and gradient levels, TSEP-Attack attains an FSIM mean of 0.710, notably superior to PGD (0.568), suggesting its enhanced capability in preserving low-level visual features such as edges and contours. Its GMSD mean of 0.236 is lower than those of PGD (0.280) and ShadowAttack (0.253), demonstrating that the introduced perturbations have minimal impact on the gradient structure and yield more natural visual appearances. At the pixel level, TSEP-Attack achieves an MAE mean of 6.322, substantially lower than PGD (10.409) and ShadowAttack (19.133), indicating more restrained pixel-wise modifications and effective control over the overall image discrepancy.

In summary, TSEP-Attack achieves remarkable visual stealthiness in generating adversarial examples, exhibiting outstanding performance across multiple dimensions including structural, feature-level, gradient-based, and pixel-wise assessments.

4.3.2. Transferability experiments

This work uses the four recognition networks as source models for generating adversarial samples. For each source model, images that were successfully attacked are sampled, and their attack performance is evaluated across all four models. This procedure verifies both the repeatability of the attacks and their transferability to different architectures.

In the attack performance experiments, 10 successful adversarial samples from the final epochs were selected per traffic sign category for each target model. The validation was performed by evaluating their attack success rates across all four victim models.

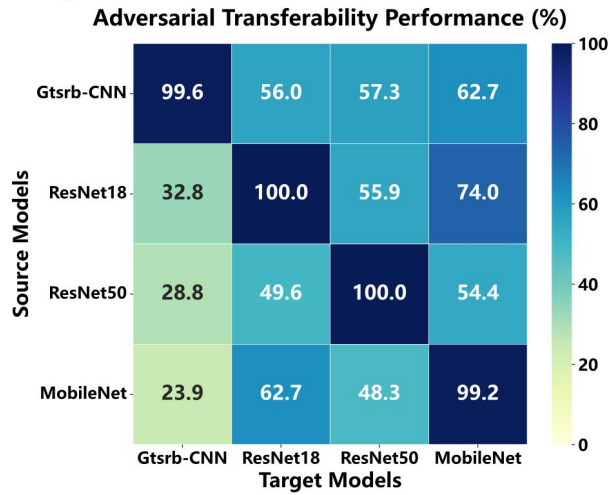

**Fig. 8.** Transferability performance of TSEP-Attack.

Figure 8 presents the results of our transferability experiments. The findings demonstrate that our method exhibits notable repeatability and considerable transferability when generating adversarial patches in a white-box setting. Although no specific transferability-enhancing modifications were implemented, the adversarial examples generated with each of the four networks as source models achieved high attack success rates against all target models except Gtsrb-CNN, reaching up to 74.0% while maintaining nearly perfect repeatability in white-box scenarios. However, adversarial examples crafted from ResNet18, ResNet50, and MobileNet showed limited transferability to Gtsrb-CNN, potentially due to architectural differences between these networks. Overall, our attack method demonstrates promising transfer capabilities and shows potential for generalized attack applicability in black-box scenarios.

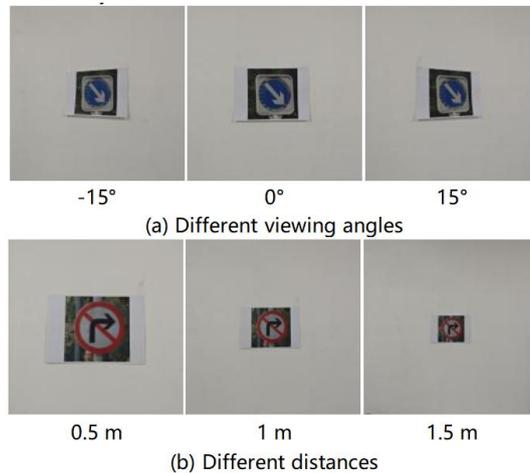

**Fig.9.** Sampled images in physical experiments at various viewpoints angles (a) and distances (b)

### 4.3.3. Physical attack test

To validate the effectiveness of the proposed method, adversarial samples with the strongest transferability performance from the Gtsrb-CNN model were printed and captured using a camera from various distances and perspectives to verify real-world attack efficacy and transfer capability. Specifically, for each category of traffic signs in the dataset, 18 images were collected under daylight conditions at shooting distances of 0.5m, 1m, and 1.5m, with viewpoints including 0° and ±15°. We first verified the recognition accuracy of original images under these shooting configurations—all clean images corresponding to the adversarial samples were correctly classified by all four models, ensuring that potential robustness issues in the recognition models themselves would not compromise experimental validity.

Figure 10 demonstrates the cropped original samples and their adversarial counterparts with recognition results, captured at 0.5m distance and 0° viewing angle, for both the Gtsrb-CNN model.

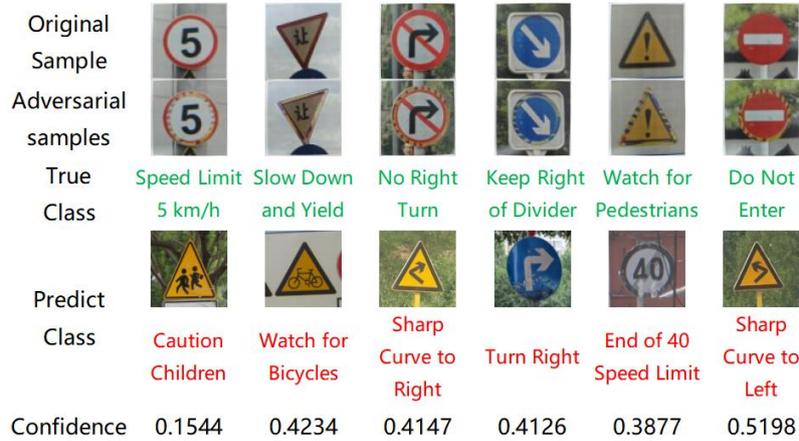

**Fig. 10.** Cropped photos of printed original and adversarial samples.

**Table 3 - Attack performance in physical domain**

| Models | Metrics | Gtsrb-CNN | MobileNet | ResNet18 | ResNet50 |
|---|---|---|---|---|---|
| Gtsrb-CNN samples | ASR(%) | 78.3 | 43.3 | 46.7 | 51.7 |
| | Confidence drop | 0.343 | 0.263 | 0.224 | 0.267 |

Table 3 presents the comprehensive ASR of adversarial examples generated by TSEP-Attack on Gtsrb-CNN model under real-world conditions across varying distances and viewing angles, along with the confidence score reduction, relative to 1.00, for failed attack samples, across different recognition models. The ASR performance in physical experiments generally aligns with prior experimental results. Although the ASR of adversarial samples decreased, their transferability remained consistent. This indicates that our method can simulate real-world sampling to a certain extent and maintains viable attack capabilities in physical environments. Specifically, the approach achieves an attack success rate of close to 80% in white-box settings while retaining a certain level of transferability in black-box scenarios.

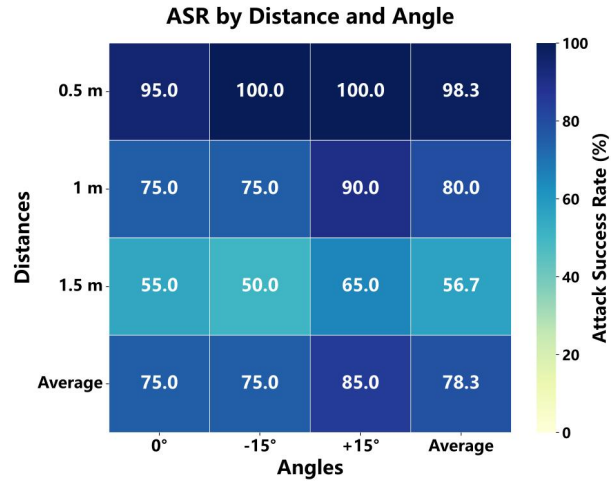

**Fig. 11.** ASR across varying physical conditions.

Figure 11 presents the white-box ASR of TSEP-Attack against the Gtsrb-CNN model under varying physical conditions. Experimental results demonstrate that while the adversarial samples generated by TSEP-Attack maintain comparable ASR across different viewing angles, the success rate decreases with increasing attack distance. It should be noted that due to the use of A4 paper for printing, the physical size of the adversarial samples was significantly reduced compared to real traffic signs. Although high-resolution sampling was employed, the variation in sampling quality caused by increasing distance appears to have affected attack effectiveness to some extent. Overall, our method demonstrates certain robustness against variations in both distance and viewing angle in real-world scenarios.

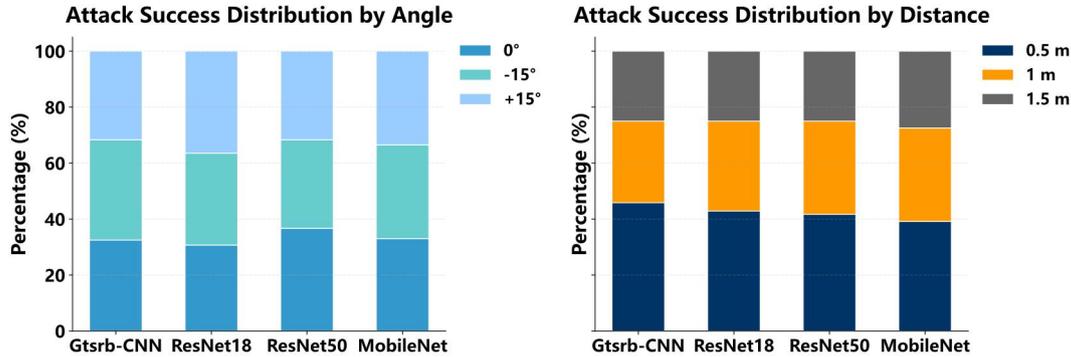

**Fig. 12.** ASR distribution of distances and viewing angles.

Figure 12 further presents the distribution of successful attack rates for physical adversarial samples generated against Gtsrb-CNN across different models under varying distances and viewing angles. The results reveal that all models maintain remarkably consistent distributions of vulnerable samples, demonstrating the spatial effectiveness consistency of our adversarial samples across different architectures.

*4.4.Ablation study*

In order to evaluate the impact of critical parameters and design configurations in our proposed method, this paper performed comprehensive ablation experiments focusing on patch width and perturbation strength.

1. Effect of patch width

The width of the annular patch mask directly determines the occlusion area and thus has a significant potential impact on attack performance. This paper evaluates the effect of three annular edge mask width ratio $r$ = 0.02, 0.04 and 0.06 on the ResNet18 model in the digital domain, as illustrated in Figure 13. The experimental results along with illustrative examples are shown in Figure 14.

The experimental results demonstrate a positive correlation between annular mask width and ASR when other parameters remain constant. Nevertheless, excessive expansion of the mask dimension results in substantial occlusion of critical sign content, thereby compromising the stealth requirement for practical adversarial attacks. Based on this trade-off analysis, a balanced configuration with the annular edge mask width ratio $r$ set to 0.04 was adopted as the default parameter for all adversarial sample generation in this study.

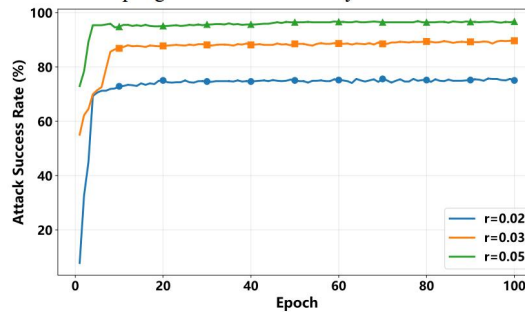

**Fig. 13.** Performance of digital patch attack of different widths.

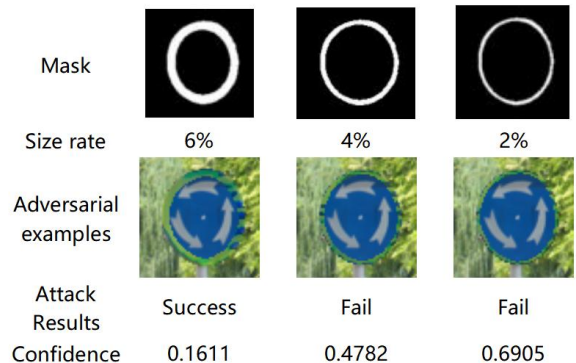

**Fig. 14.** Adversarial images and masks of different widths.

## 5. Conclusions and Future Work

The core contribution of this research lies in the development of TSEP-Attack - a physical adversarial patch method designed for traffic sign recognition systems. Through the integrated framework of edge-aware mask generation, multi-level loss function optimization, and adaptive training scheduling, our approach achieves breakthrough performance across three key dimensions: In attack effectiveness, it demonstrates exceptional performance in physical attack simulation experiments, achieving up to 91.95% success rate in white-box attacks while exhibiting remarkable cross-model transferability. Regarding stealth characteristics, the generated adversarial patches significantly outperform existing methods across multiple visual stealth metrics, achieving natural integration with environmental contexts. In terms of physical robustness, the method maintains stable attack performance under varying capture angles and distances within a comprehensive evaluation framework. These combined attributes ensure the method's practical utility and effectiveness in real-world scenarios.

Future research will develop along three main directions: First, we will validate the attack capability of our method against broader perception tasks—including detection, recognition, and classification of traffic signs—under more complex and diverse physical environments to enhance its generalization ability. Second, we plan to evaluate the method across more datasets and model architectures, conducting extensive and fair comparisons with a wider range of physical adversarial attacks to thoroughly assess its performance. Finally, we will carry out human perceptual studies to rigorously evaluate the stealth of TSEP-Attack and further refine our concealment strategies.

## Declaration of Competing Interest

The authors declare that they have no known competing financial interests or personal relationships that could have appeared to influence the work reported in this paper.

## CRediT Authorship Contribution Statement

**Haojie Ji:** Conceptualization, Methodology. **Te Hu:** software, writing - original draft. **Haowen Li:** Investigation, Formal analysis. **Long Jin:** Data curation. **Chongshi Xin:** Writing - review & editing. **Yuchi Yao:** Validation. **Jiarui Xiao:** Visualization.

## Acknowledgment

This work was supported by the Beijing Municipal Education Commission Science and Technology Program (Grant No. KM 202411232005).